\documentclass{article}
\usepackage{spconf,amsmath,amssymb,graphicx,hyperref,bm}
\usepackage{tikz}
\usepackage{pgfplots}
\pgfplotsset{compat=1.18}
\usepgfplotslibrary{groupplots}
\usepackage{xcolor}
\usepackage{subcaption}
\usepackage{algorithm}
\usepackage{algorithmic}
\usepackage{float}
\usepackage{tcolorbox}
\usepackage{array}
\usepackage{colortbl}
\usetikzlibrary{positioning}  % Add this line
\definecolor{sendercolor}{RGB}{240,248,255}    % Alice blue for sender
\definecolor{receivercolor}{RGB}{245,255,245}  % Honeydew for receiver
\definecolor{headercolor}{RGB}{70,130,180}     % Steel blue for headers
\definecolor{bordercolor}{RGB}{100,149,237}    % Cornflower blue for borders

\definecolor{lightblue}{RGB}{173,216,230}      % Code 0
\definecolor{lightgreen}{RGB}{144,238,144}     % Code 1
\definecolor{orange}{RGB}{255,165,0}           % Code 2
\definecolor{purple}{RGB}{128,0,128}           % Code 3

\title{Residual Vector Quantization for Communication-Efficient Multi-Agent Perception}
\name{Dereje Shenkut \qquad B.V.K. Vijaya Kumar}
\address{Carnegie Mellon University}
\begin{document}
\ninept

\maketitle
\begin{abstract}
Multi-agent collaborative perception (CP) improves scene understanding by sharing information across connected agents such as autonomous vehicles, unmanned aerial vehicles, and robots. Communication bandwidth, however, constrains scalability. We present ReVQom, a learned feature codec that preserves spatial identity while compressing intermediate features. ReVQom is an end-to-end method that compresses feature dimensions via a simple bottleneck network followed by multi-stage residual vector quantization (RVQ). This allows only per-pixel code indices to be transmitted, reducing payloads from $8192$ bits per pixel (bpp) of uncompressed 32-bit float features to $6$–$30$ bpp per agent with minimal accuracy loss. On DAIR-V2X real-world CP dataset, ReVQom achieves $273\times$ compression at 30 bpp to $1365\times$ compression at 6 bpp. At $18$ bpp ($455\times$), ReVQom matches or outperforms raw-feature CP, and at $6$–$12$ bpp it enables ultra-low-bandwidth operation with graceful degradation. ReVQom allows efficient and accurate multi-agent collaborative perception with a step toward practical V2X deployment.
\end{abstract}
%
% \begin{keywords}
%   Multi-agent collaborative perception, V2X communication, residual vector quantization, feature compression, bandwidth-efficient sensing, autonomous driving
% \end{keywords}
%
\section{Introduction}
\label{sec:intro}

  \begin{figure*}[htb]
    \input{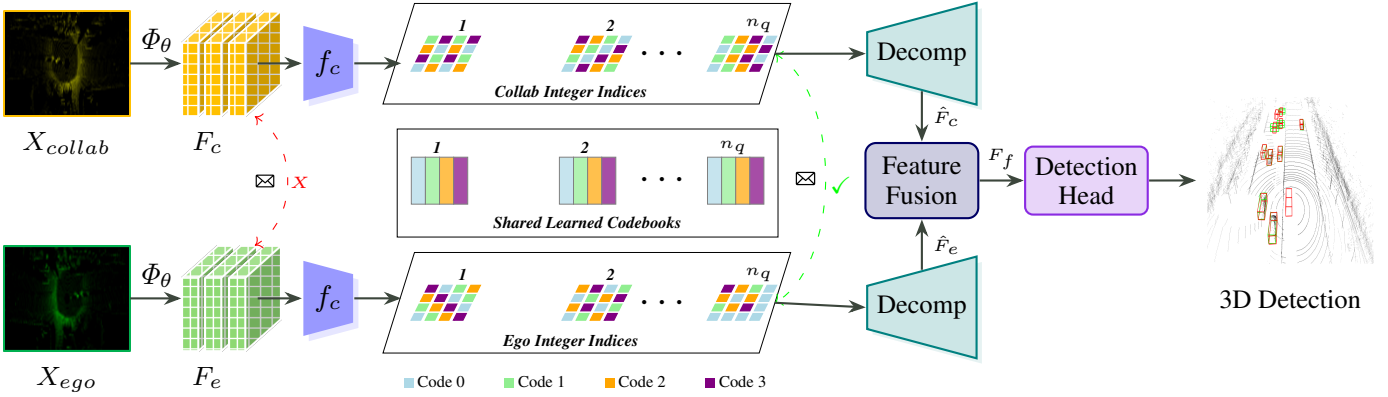}
    \caption{Each agent first extracts BEV features with a sparse voxel encoder ($\phi_\theta$) and then applies a $1 \times 1$ bottleneck represented by $f_c$ and $n_q$-stage residual vector quantization (RVQ) to produce per-pixel code indices. Transmission occurs using only indices ($\approx HW n_q \log_2 K$ bits) as opposed to full features ($32 \times C \times H \times W$ bits) where $C$ is the number of channels, $H$ is the height, and $W$ is the width of the feature map and $K$ is the codebook size in a stage. With a shared codebook, the receiver first decodes the indices and then reconstructs the full feature $\hat{F}$ with decompressor ($Decomp$). $F_c$ and $F_e$ denote raw BEV features from ego agent and collaborator agent respectively, with $\hat{F}_c$ and $\hat{F}_e$ being their reconstructed versions, and $F_f$ the final fused features. This allows an integer indices based communication between agents providing compression without overly compromising spatial fidelity.}
    \label{fig:remap_overview}
    \end{figure*}
Accurate perception and environmental understanding form the foundation of autonomous vehicle systems. Learning-based multi-sensor fusion with cameras, LiDAR, and radar \cite{Geiger2012CVPR, nuscenes, waymo2020} has been extensively studied, yet long-range detection and occlusion remain challenging because all sensors are mounted on a single vehicle which limits sensing range and robustness to occlusions. Multi-agent collaborative perception (CP) addresses these limitations by allowing vehicles and infrastructure units to exchange information over V2X communications \cite{wang2020v2vnet, xu2022opencood}. Existing CP approaches focused on three kinds of fusion strategies. Early fusion exchanges raw sensor data \cite{Li_2021_NeurIPS}, intermediate fusion shares learned feature representations \cite{xu2022v2xvit, CoBEVT}, and late fusion transmits only detection outputs \cite{cui2022coopernaut}. Recent advances have introduced vision transformer-based collaboration \cite{xu2022v2xvit}, robust asynchronous perception \cite{lei2022latencyaware}, and heterogeneous multi-agent frameworks \cite{HEAL, xu2023bridging}. 

Communication bandwidth constraints remain a critical bottleneck for practical deployment and scalability to multiple agents. Current approaches achieve limited compression ratios. \cite{hu2022where2comm} uses spatial attention for selective transmission, while \cite{what2comm} applies feature decoupling. Recent work explores codebook-based compression \cite{CodeFilling} and point cluster representations \cite{CPPC}, but these methods either compromise spatial geometry or achieve insufficient compression for bandwidth-constrained V2X scenarios.
A fundamental limitation is that existing methods focus on \emph{what} features to transmit rather than \emph{how} to compress them aggressively while preserving spatial structure. To that end, we propose ReVQom shown in Fig. \ref{fig:remap_overview}, a method that enables aggressive compression with minimal performance loss. ReVQom first applies a spatially identity-preserving $1\times 1$ convolution with feature normalization to reduce the number of channels, then performs multi-stage residual vector quantization (RVQ) \cite{RVQ}. In this setup, only per-pixel codebook indices are transmitted, and the receiver reconstructs features using pre-shared exponential moving average (EMA) updated codebooks. This maintains the geometry required for accurate bird's eye view (BEV) fusion. Our contributions are threefold: (1) a spatial-identity-preserving codec with pre-shared codebooks and an index-only messaging scheme, (2) a practical integration with multi-agent BEV fusion that operates under tight bandwidth budgets, and (3) a systematic ablation study that highlights the benefit of faster codebook adaptation and compact quantization settings that balance rate and detection quality.

\section{Related Work}

\label{sec:related}
\textbf{Collaborative Perception.} Collaborative perception (CP) via V2X communications enhances autonomous vehicles' ability to perceive through occlusions and extend detection ranges \cite{xu2022v2xvit}. Among the three fusion techniques discussed in Section~\ref{sec:intro},
V2VNet \cite{wang2020v2vnet} introduces intermediate-level fusion where vehicles compress and exchange network representations via graph neural networks to improve single agent perception and prediction performance. Another notable work, DiscoNet \cite{Li_2021_NeurIPS} employs knowledge distillation-based fusion with teacher models using early-fusion and students using graph-based intermediate fusion. AttFuse \cite{xu2022opencood} introduces attention-based V2V collaboration with the widely adopted OPV2V benchmark. In tackling practical CP non-idealistic conditions, V2X-ViT \cite{xu2022v2xvit} introduces vision transformer-based collaboration robust to noisy localization, while SyncNet \cite{lei2022latencyaware} addresses latency-aware collaboration through temporal alignment. CoAlign \cite{CoAlign} introduces another imperfect localization mitigation technique specifically targeting unknown pose errorsusing pose graph modelling. HEAL \cite{HEAL},  Hetecooper \cite{Hetecooper}, GenComm \cite{GenComm} and NegoCollab \cite{NegoCollab} introduce multi-agent feature cooperative method to handle heterogeneous agents and sensor modalities. \\
\textbf{Information Selection and Compression in Collaborative Perception.} Communication constraints are crucial for practical CP deployment \cite{shenkut2024impact}. Where2comm \cite{hu2022where2comm} optimizes bandwidth usage by selectively transmitting informative features achieving per agent compression. Recent compression advances include CodeFilling \cite{CodeFilling}, which uses information filling with codebooks. CPPC \cite{CPPC} introduces point cluster representations for compact messaging.  FocalComm \cite{shenkut2025focalcomm} introduces difficulty-aware agent selection by mining hard instances for targeted collaboration. However, current CP methods focus on optimizing what features to transmit rather than how to compress them efficiently. Existing compression approaches use channel attention \cite{hu2022where2comm} or spatial attention \cite{lei2022latencyaware} to reduce feature dimensions, but achieve limited compression ratios. 
\begin{table}[thb]
  \centering
  \caption{RVQ Based Collaboration Protocol}
  \small
  \begin{tabular}{p{0.45\textwidth}}
  \hline
  \textbf{Sender Agent (Encoding)} \\
  \hline
  $\mathbf{F} \in \mathbb{R}^{H \times W \times C}$ $\triangleright$ \textit{input BEV features} \\
  $\mathbf{F}_r = \text{GroupNorm}(\text{Conv}_{1 \times 1}(\mathbf{F})) \in \mathbb{R}^{H \times W \times C_r}, \quad C_r \ll C$ $\triangleright$ \textit{channel reduction} \\[0.2cm]

  $\mathbf{r}^{(0)} = \mathbf{F}_r$ $\triangleright$ \textit{initialize residual} \\
  For $i = 0, 1, \ldots, n_q-1$: \\
  \quad $k^{(i)} = \arg\min_{k \in \{1,\ldots,K\}} \|\mathbf{r}^{(i)} - \mathbf{e}_k^{(i)}\|_2^2$ $\triangleright$ \textit{nearest codebook} \\
  \quad $\mathbf{q}^{(i)} = \mathbf{e}_{k^{(i)}}^{(i)}$ $\triangleright$ \textit{quantized vector} \\
  \quad $\mathbf{r}^{(i+1)} = \mathbf{r}^{(i)} - \mathbf{q}^{(i)}$ $\triangleright$ \textit{update residual} \\
  \quad $\mathbf{e}_{k^{(i)}}^{(i)} \leftarrow (1-\alpha)\mathbf{e}_{k^{(i)}}^{(i)} + \alpha\mathbf{r}^{(i)}$ $\triangleright$ \textit{EMA update} \\[0.2cm]

  $\mathcal{I} = \{k^{(0)}, k^{(1)}, \ldots, k^{(n_q-1)}\}$ $\triangleright$ \textit{transmit indices} \\
  $R = n_q \log_2 K$ bits per spatial location $\triangleright$ \textit{bitrate}
  \\[0.3cm]
  \hline
  \textbf{Receiver Agent (Decoding)} \\
  \hline
  $\mathcal{I} = \{k^{(0)}, k^{(1)}, \ldots, k^{(n_q-1)}\}$ $\triangleright$ \textit{receive indices} \\
  $\mathbf{q}^{(i)} = \mathbf{e}_{k^{(i)}}^{(i)}$ for $i = 0, 1, \ldots, n_q-1$ $\triangleright$ \textit{codebook lookup} \\[0.2cm]

  $\mathbf{z}_q = \sum_{i=0}^{n_q-1} \mathbf{q}^{(i)}$ $\triangleright$ \textit{accumulate quantized vectors} \\
  $\mathbf{z}'_q = \text{ReLU}(\text{Conv}_{1 \times 1}(\mathbf{z}_q))$ $\triangleright$ \textit{post-affine transformation} \\
  $\hat{\mathbf{F}} = \text{ReLU}(\text{Conv}_{1 \times 1}(\text{GroupNorm}(\mathbf{z}'_q)))$ $\triangleright$ \textit{Channel Expansion} \\[0.2cm]

  $\mathbf{F}_{\text{f}} = \gamma(\hat{\mathbf{F}}, \mathbf{F}_{\text{local}})$ $\triangleright$ \textit{multi-agent fusion}
  \\
  \hline
  \end{tabular}
  \label{tab:communication_protocol}
\end{table}
\section{Residual Vector Quantization for Multi-Agent Perception}
\subsection{Multi-Agent Perception}
In multi-agent collaborative perception, our goal is to achieve bandwidth-efficient transmission while maximizing perception performance for all agents in a distributed setting or for the ego agent in a centralized setting. Consider a scenario with $N$ collaborative agents, where agent $i$ observes sensor data $\mathbf{X}_i$ with corresponding ground-truth annotations $\mathbf{Y}_i$. The objective of collaborative perception is to maximize detection performance across all agents subject to a communication budget $B$. It is formulated as:
\begin{equation}
\begin{aligned}
\max_{\theta, \{\mathbf{m}_{ij}\}} &\sum_{i=1}^N \mathcal{P}_{\text{det}}\left(f_\theta\left(\mathbf{X}_i, \{\mathbf{m}_{ji}\}_{j \in \mathcal{N}_i}\right), \mathbf{Y}_i\right) \\
&\text{subject to } \sum_{i=1}^N \sum_{j \in \mathcal{N}_i} \|\mathbf{m}_{ij}\| \leq B
\end{aligned}
\end{equation}
where $f_\theta(\cdot)$ is the detection network parameterized by $\theta$, $\mathcal{P}_{\text{det}}(\cdot, \cdot)$ is the detection performance metric (e.g., mAP), $\mathbf{m}_{ij} \in \mathbb{R}^{C \times H \times W}$ denotes the message transmitted from agent $i$ to agent $j$, $\mathcal{N}_i$ represents the neighborhood set of agent $i$, and $\|\mathbf{m}_{ij}\|$ measures the communication cost of message $\mathbf{m}_{ij}$. In intermediate fusion approaches, messages $\mathbf{m}_{ij}$ typically consist of Bird's-Eye-View (BEV) feature representations $\mathbf{F}_i \in \mathbb{R}^{C \times H \times W}$ extracted by agent $i$'s encoder $\phi_\theta$ such that $\mathbf{F}_i = \phi_\theta(\mathbf{X}_i)$.

\subsection{Multi-Agent Message Exchange Protocol Via Residual Vector Quantization}
Given high-dimensional BEV features $\mathbf{F} \in \mathbb{R}^{H \times W \times C}$ per agent, we achieve spatially-identity preserving compression through channel reduction followed by residual vector quantization as detailed in Table~\ref{tab:communication_protocol}. The sender first reduces channel dimensionality via $1 \times 1$ convolution and applies normalization $\mathbf{F}_r = \text{GroupNorm}(\text{Conv}_{1 \times 1}(\mathbf{F})) \in \mathbb{R}^{H \times W \times C_r}$, where $C_r \ll C$ with channel reduction ratio $C_{rr}$. We then apply $n_q$-stage RVQ with $\ell_2$ distance-based similarity matching. For each spatial position and quantization stage $i \in \{0, 1, \ldots, n_q-1\}$, we find the nearest codebook entry $k^{(i)} = \arg\min_{k \in \{1,\ldots,K\}} \|\mathbf{r}^{(i)} - \mathbf{e}_k^{(i)}\|_2^2$, where $\mathbf{r}^{(0)} = \mathbf{F}_r$ initializes the residual. The quantized vector $\mathbf{q}^{(i)} = \mathbf{e}_{k^{(i)}}^{(i)}$ is subtracted to update the residual $\mathbf{r}^{(i+1)} = \mathbf{r}^{(i)} - \mathbf{q}^{(i)}$. Codebooks $\mathbf{E}_i \in \mathbb{R}^{K \times C_r}$ are updated via EMA as $\mathbf{e}_{k^{(i)}}^{(i)} \leftarrow (1-\alpha)\mathbf{e}_{k^{(i)}}^{(i)} + \alpha\mathbf{r}^{(i)}$, where $\alpha$ is EMA decay rate. Codebooks $\mathbf{e}_k^{(i)}$ are learned during training and stored identically on all agents, with the EMA parameter $\alpha$ preset across all agents to ensure synchronized codebook updates.  This enables only indices $\mathcal{I} = \{k^{(0)}, k^{(1)}, \ldots, k^{(n_q-1)}\}$ to be transmitted, requiring $R = n_q \log_2 K$ bits per spatial location. The receiver reconstructs features by accumulating codebook lookups $\mathbf{z}_q = \sum_{i=0}^{n_q-1} \mathbf{q}^{(i)}$, applying post-affine transformation, and channel expansion back to $C$ dimensions $\hat{\mathbf{F}} = \text{ReLU}(\text{Conv}_{1 \times 1}(\text{GroupNorm}(\mathbf{z}'_q)))$, and proceeds to multi-agent fusion $\mathbf{F}_{\text{f}} = \gamma(\hat{\mathbf{F}}, \mathbf{F}_{\text{local}})$. \\
\textbf{Loss Functions.} Our training objective combines multiple loss terms with EMA-based codebook updates:
{\setlength{\abovedisplayskip}{4pt}\setlength{\belowdisplayskip}{4pt}
\begin{align}
\mathcal{L} &= \mathcal{L}_{\text{task}} + \mathcal{L}_{\text{VQ}} + \mathcal{L}_{\text{ortho}} \\
\mathcal{L}_{\text{VQ}} &= \beta_{\text{commit}} \|\text{sg}[\mathbf{z}_q] - \mathbf{z}\|_2^2 \\
\mathcal{L}_{\text{ortho}} &= \lambda_{\text{ortho}} \|\mathbf{W}_e \mathbf{W}_e^T - \mathbf{I}_{C_r}\|_F^2
\end{align}}%
where $\mathcal{L}_{\text{task}}$ is the main detection loss (focal loss for classification and regression), $\mathcal{L}_{\text{VQ}} = \beta_{\text{commit}} \|\text{sg}[\mathbf{z}_q] - \mathbf{z}\|_2^2$ is the vector quantization commitment loss \cite{vqvae} that ensures that encoded features $\mathbf{z}$ remain close to their quantized versions $\mathbf{z}_q$ (where $\text{sg}[\cdot]$ denotes stop-gradient operation and $\beta_{\text{commit}}$ controls the commitment strength), and $\mathcal{L}_{\text{ortho}} = \lambda_{\text{ortho}} \|\mathbf{W}_e \mathbf{W}_e^T - \mathbf{I}_{C_r}\|_F^2$ enforces orthogonality constraints \cite{orthogonal_reg} on the encoder weight matrix $\mathbf{W}_e$ to improve feature disentanglement (where $\mathbf{I}_{C_r}$ is the identity matrix and $\lambda_{\text{ortho}}$ is the regularization weight).
\section{NUMERICAL EXPERIMENTAL RESULTS}
\label{sec:experiments}
\begin{figure*}[htb]
  \centering
  \begin{subfigure}{0.24\textwidth}
    \includegraphics[width=\textwidth]{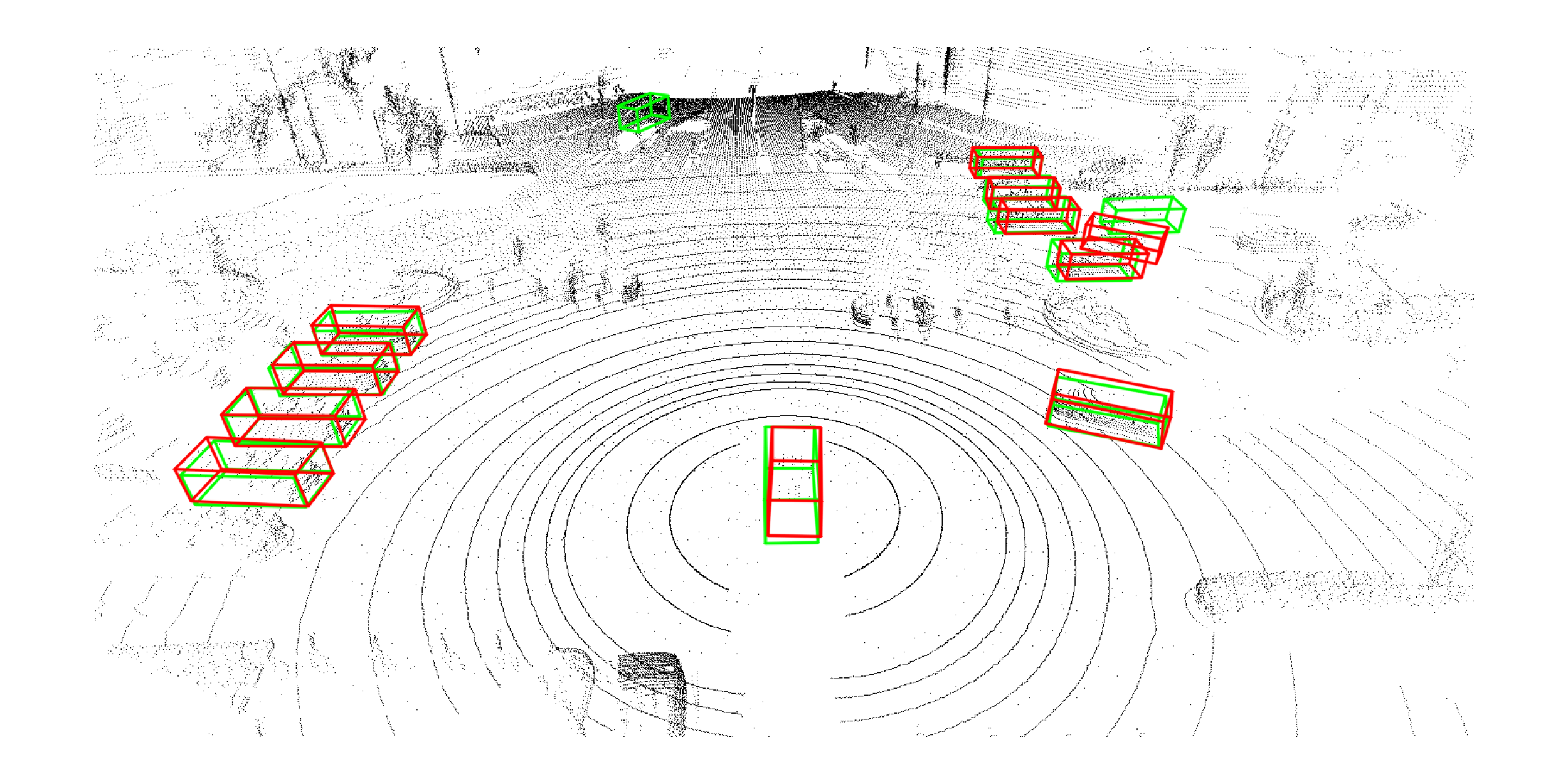}
  \end{subfigure}
  \hfill
  \begin{subfigure}{0.24\textwidth}
    \includegraphics[width=\textwidth]{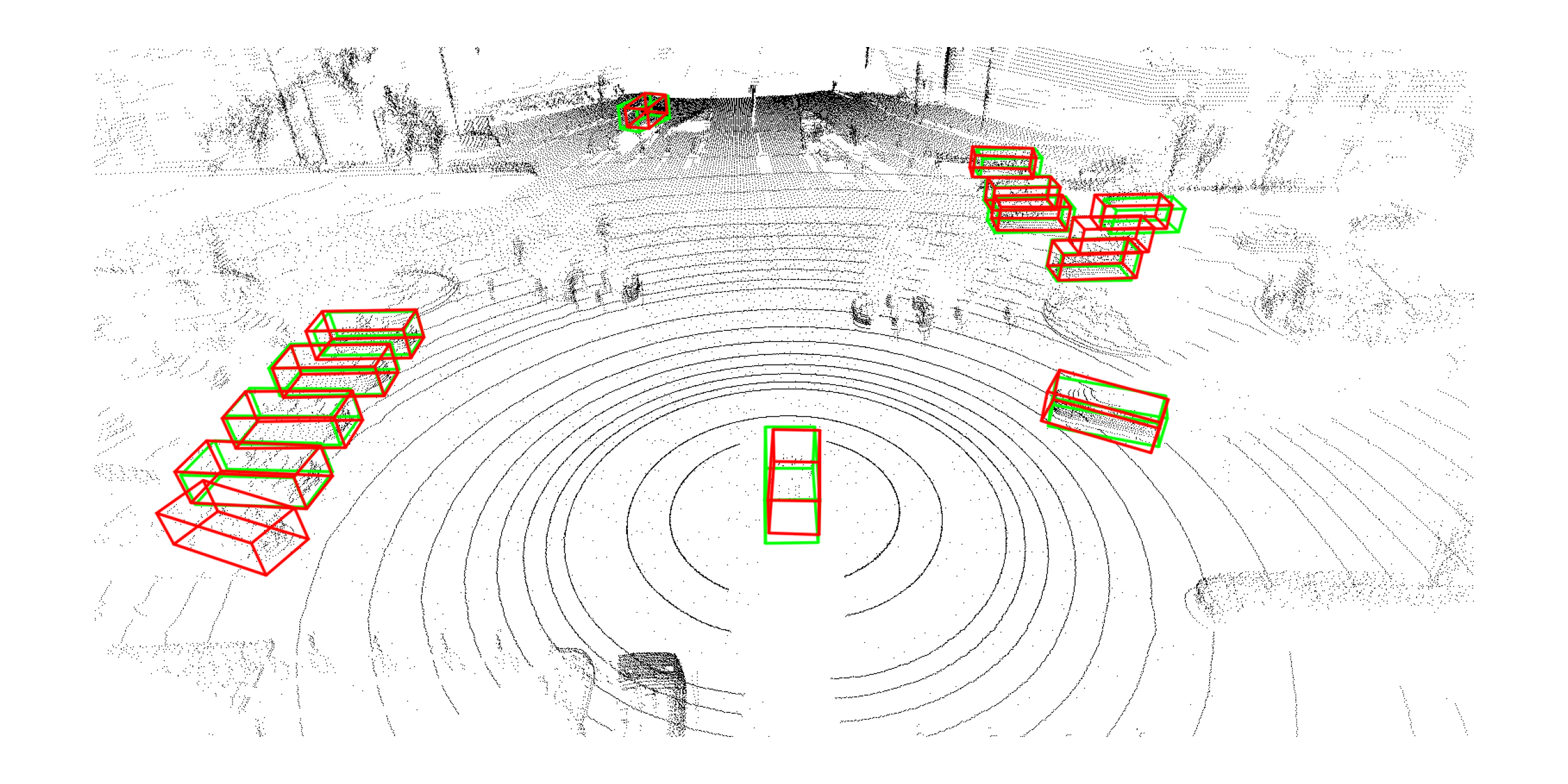}
  \end{subfigure}
  \hfill
  \begin{subfigure}{0.24\textwidth}
    \includegraphics[width=\textwidth]{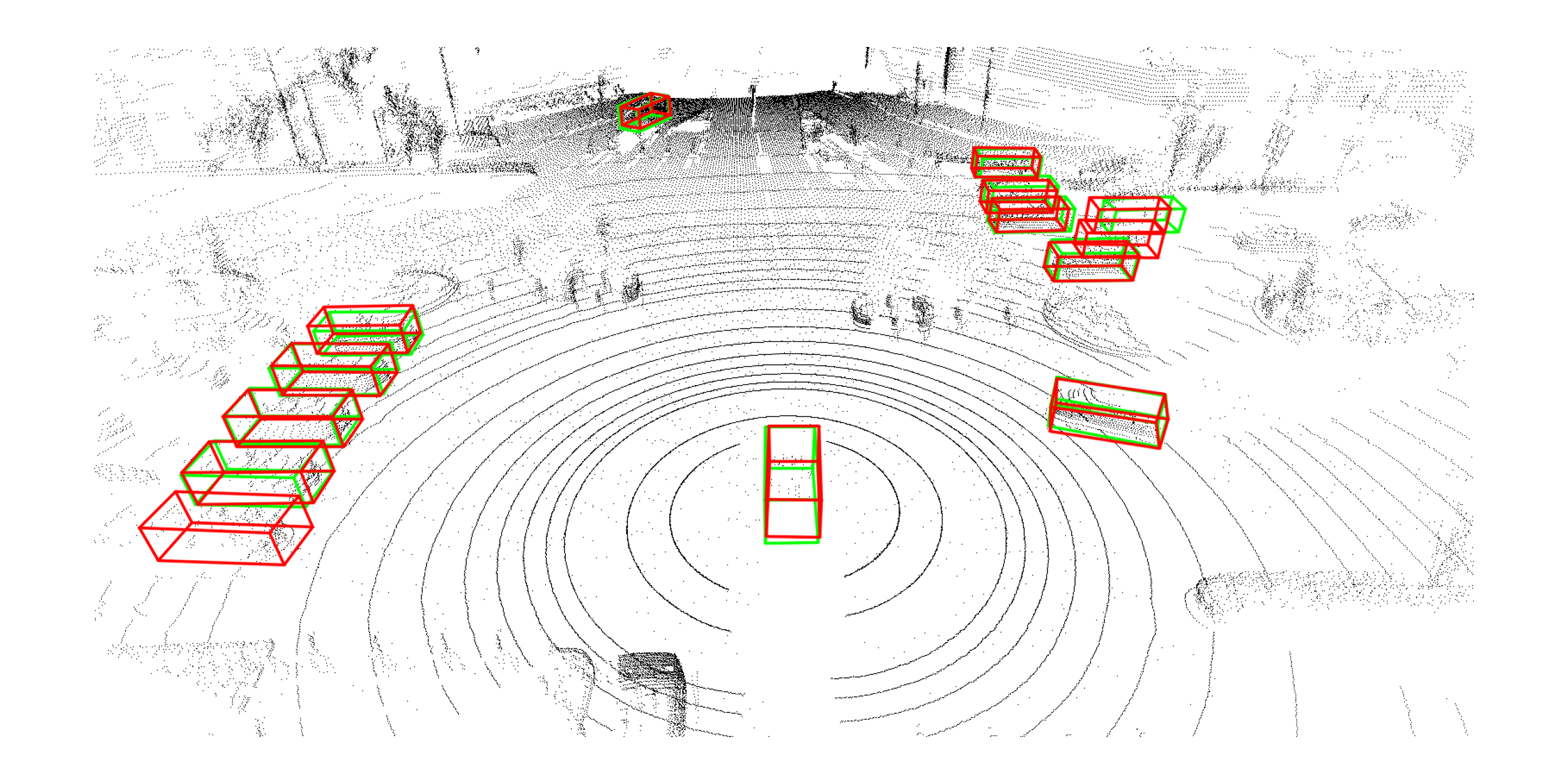}
  \end{subfigure}
  \hfill
  \begin{subfigure}{0.24\textwidth}
    \includegraphics[width=\textwidth]{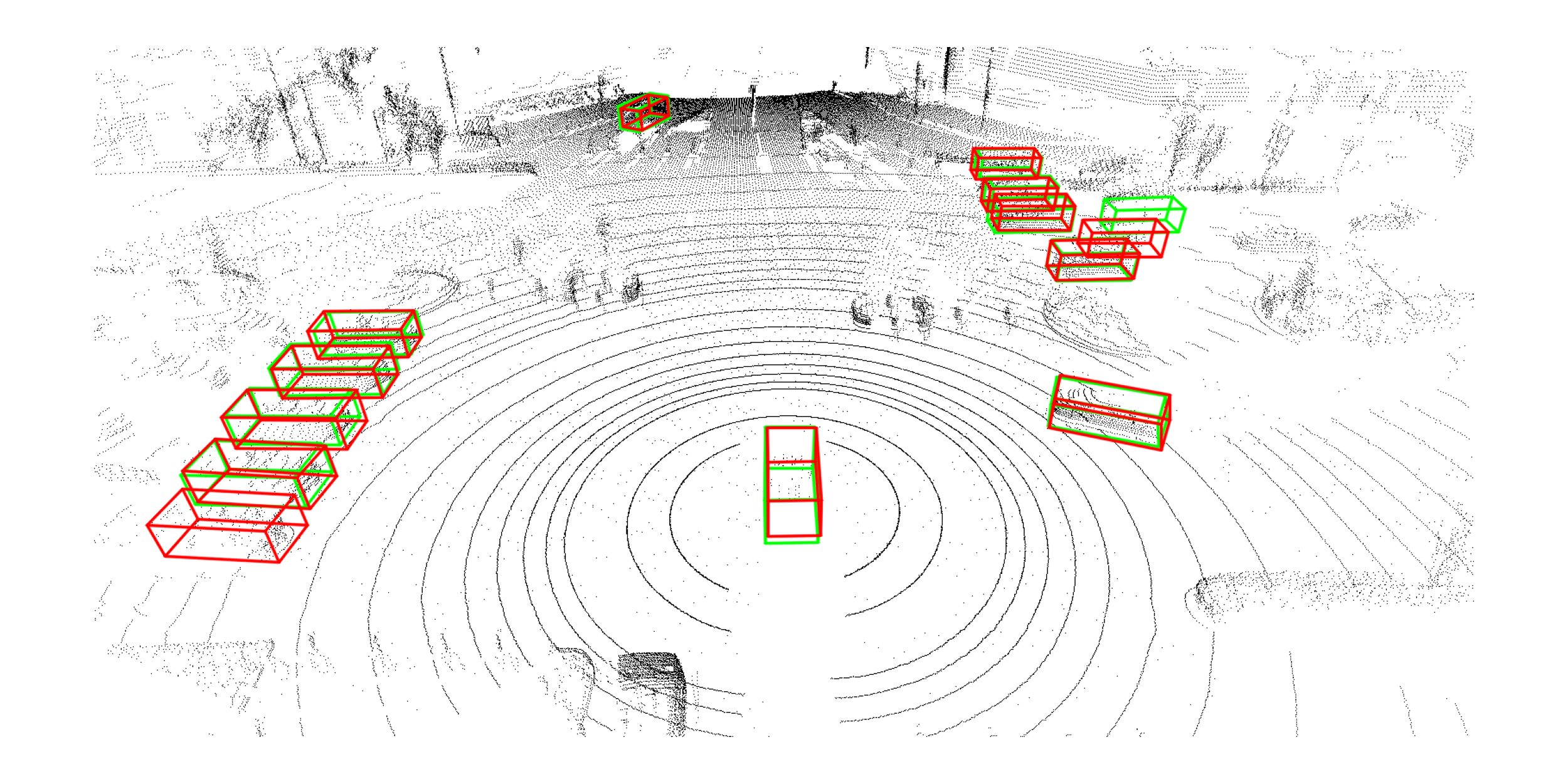}
  \end{subfigure}
  \begin{subfigure}{0.24\textwidth}
    \includegraphics[width=\textwidth]{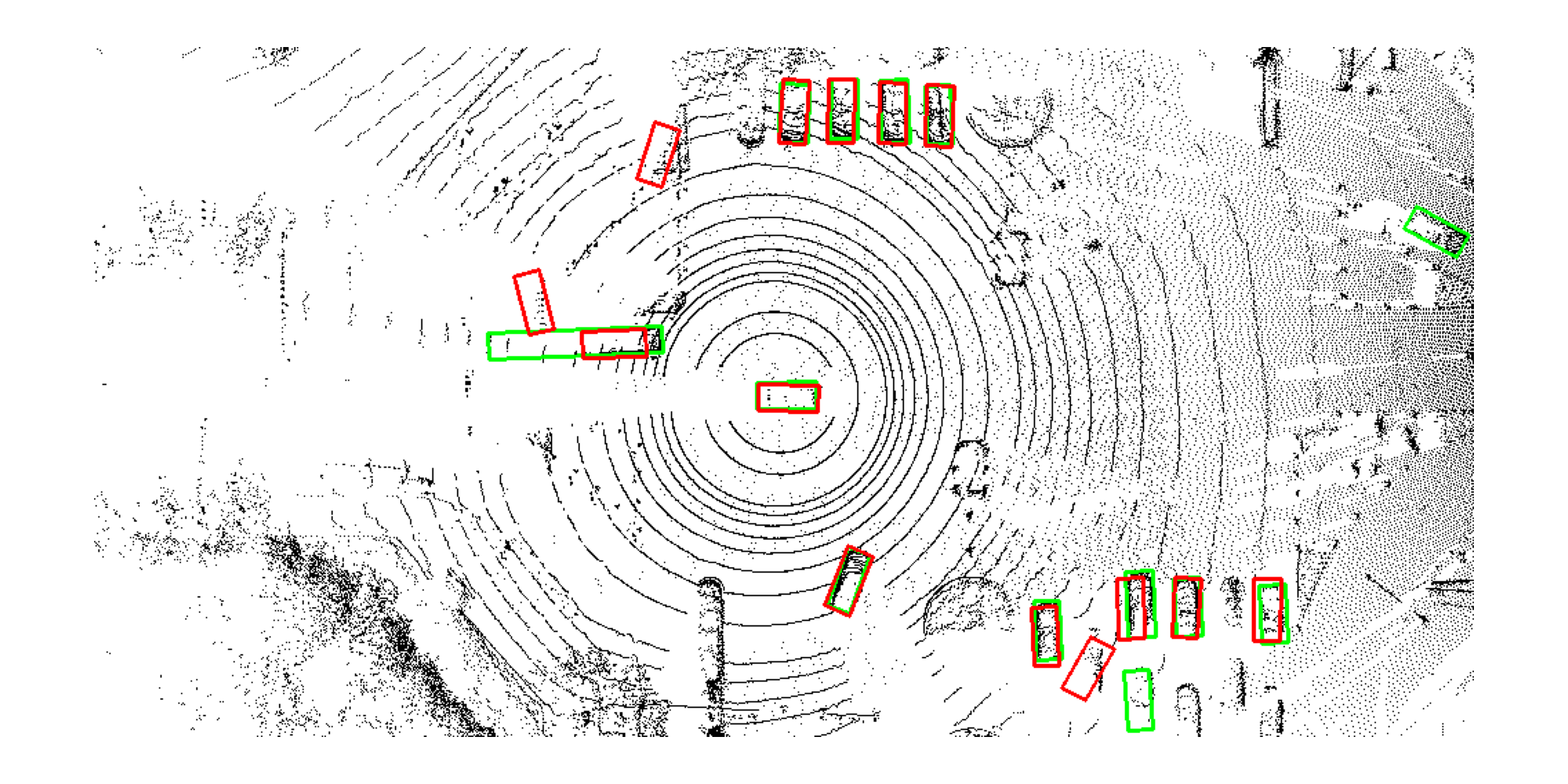}
    \caption{$6\text{ }bits$ $(K=4)$}
    \label{fig:qual_6bits}
  \end{subfigure}
  \hfill
  \begin{subfigure}{0.24\textwidth}
    \includegraphics[width=\textwidth]{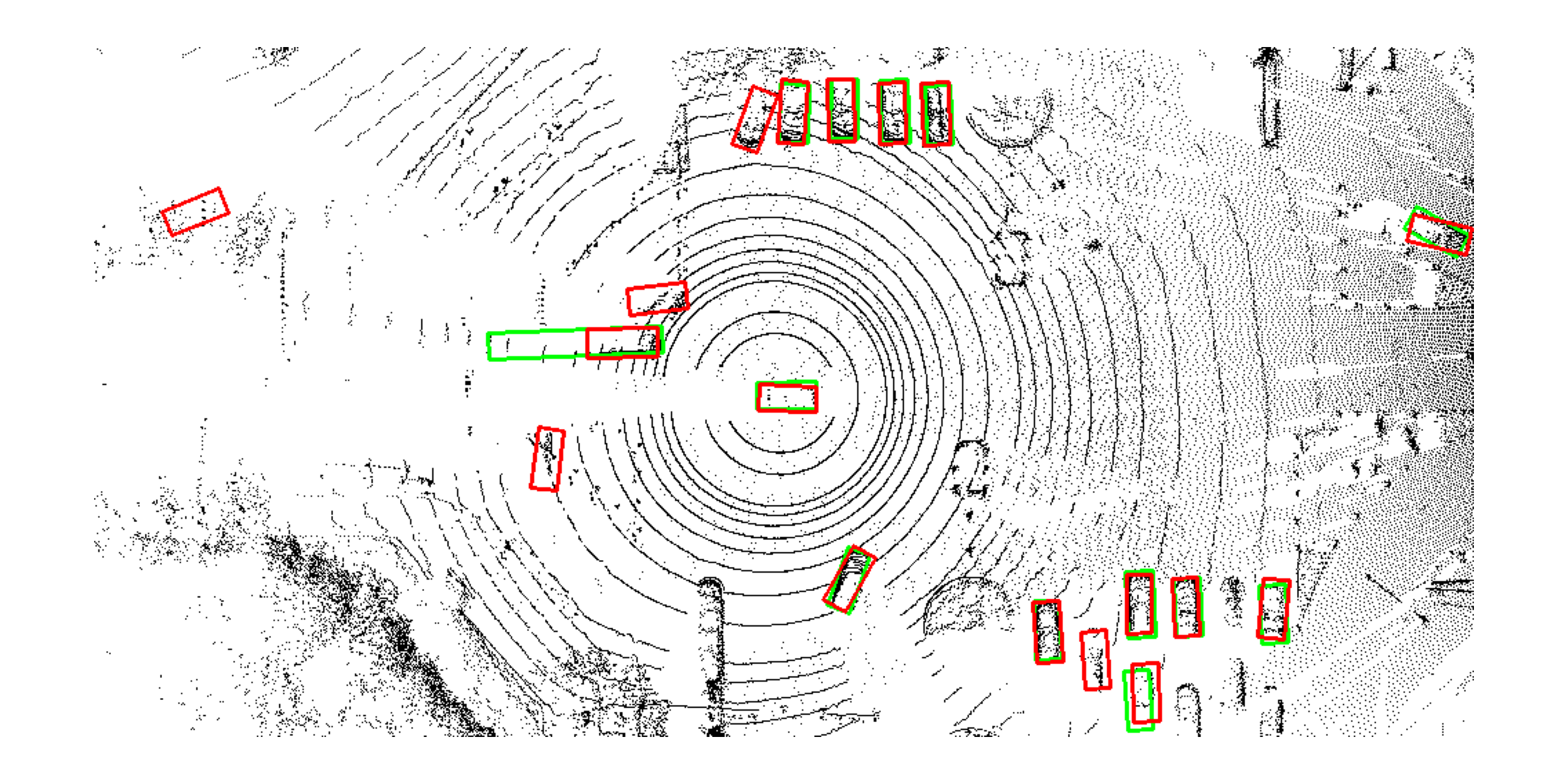}
    \caption{$18\text{ }bits$ $(K=64)$}
    \label{fig:qual_18bits}
  \end{subfigure}
  \hfill
  \begin{subfigure}{0.24\textwidth}
    \includegraphics[width=\textwidth]{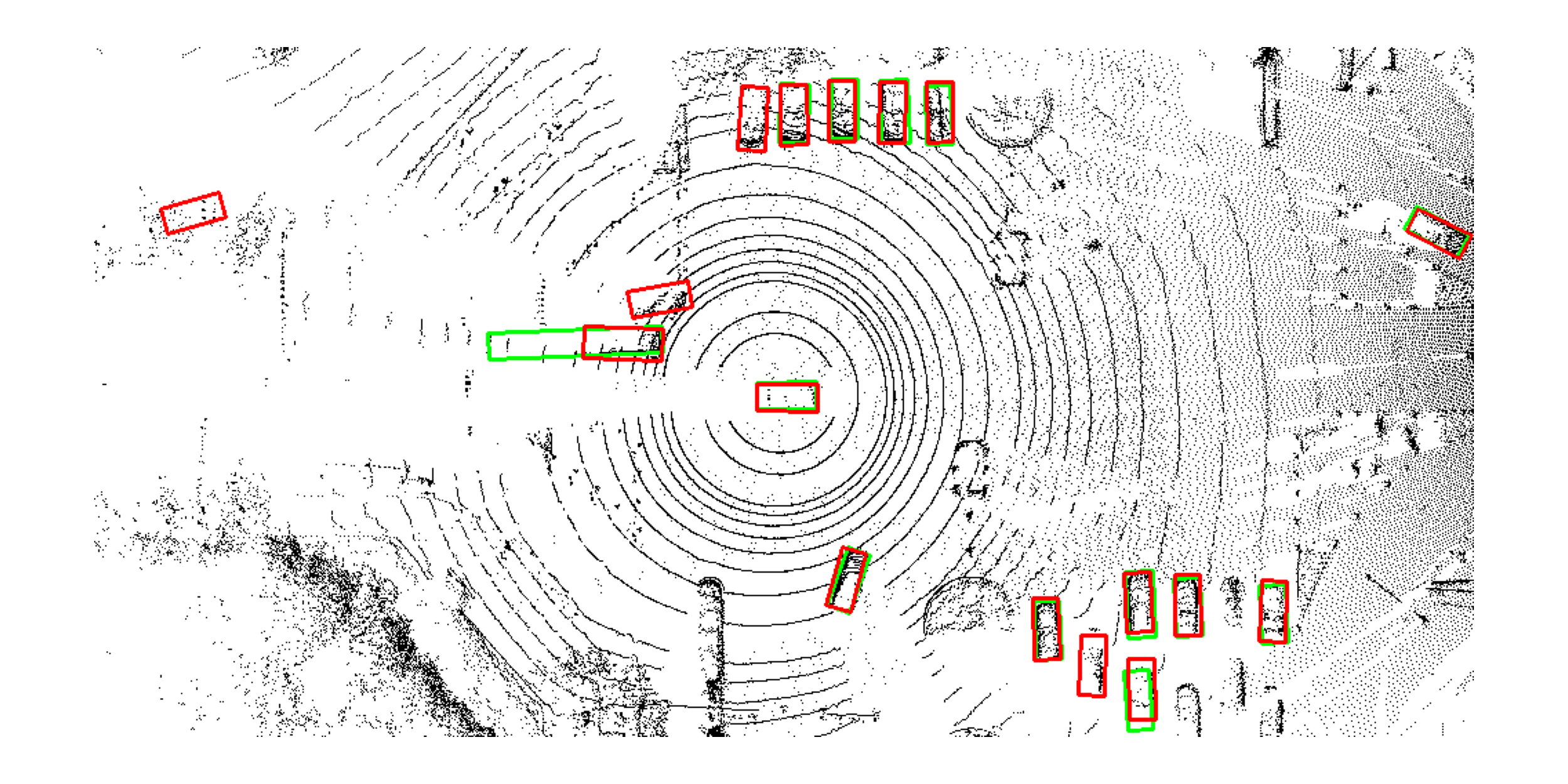}
    \caption{$24\text{ }bits$ $(K=256)$}
    \label{fig:qual_24bits}
  \end{subfigure}
  \hfill
  \begin{subfigure}{0.24\textwidth}
    \includegraphics[width=\textwidth]{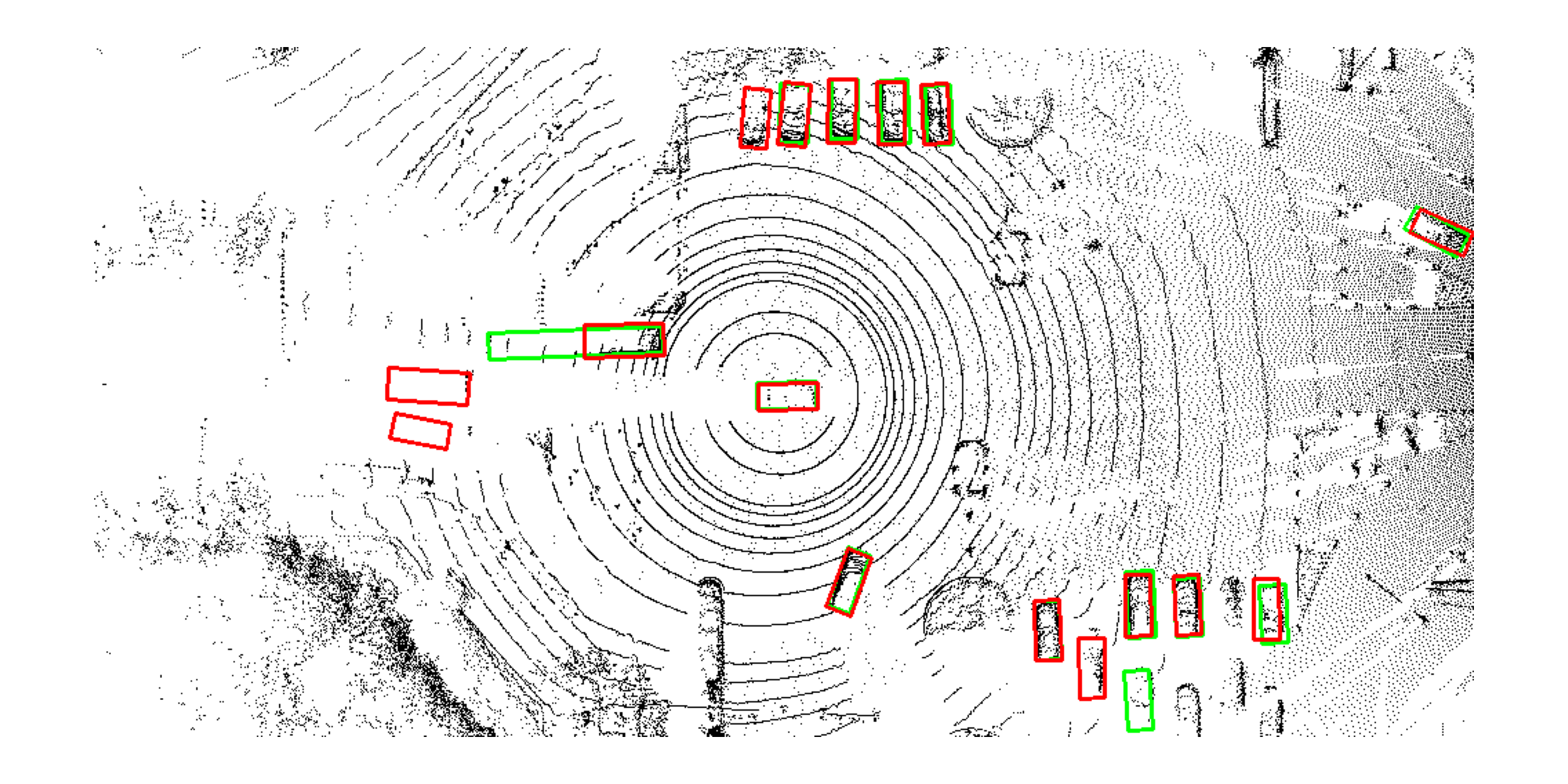}
    \caption{$8192\text{ }bits$ (Raw)}
    \label{fig:qual_raw}
  \end{subfigure}
  \caption{Detection results showing improvement with increased bit rate until optimal codebook size. The top row shows 3D view from ego vehicle's forward view perspective. The bottom row shows Bird's-eye view of the same scene with the ego vehicle at the center going left to right. \textcolor{green}{Green} and \textcolor{red}{red} boxes represent ground truth and predictions respectively. As little as 6 bits (a) works reasonably well, while higher bit rate shows improvement in precise bound box alignment between the ground truth and predictions.}
  \label{fig:qualitative_results}
\end{figure*}

\subsection{Dataset and Implementation}

\textbf{Dataset.} We evaluate ReVQom on DAIR-V2X \cite{yu2022dairv2x}, a real-world collaborative perception dataset for 3D object detection. DAIR-V2X features vehicle-infrastructure cooperation scenarios with synchronized LiDAR point clouds from roadside unit (RSU) and a connected autonomous vehicle (CAV). We additionally evaluate on OPV2V \cite{xu2022opencood}, an extensive simulated benchmark with diverse driving scenarios. \\
\textbf{Evaluation.} We follow standard evaluation protocols using AP@0.3 and AP@0.5 metrics for 3D vehicle detection, where AP@0.3 and AP@0.5 denote Average Precision at Intersection over Union (IoU) thresholds of 0.3 and 0.5 respectively. \\
\textbf{Implementation.} Our implementation uses CoBEVT \cite{CoBEVT} as the multi-agent fusion backbone, with ReVQom integrated as a plug-in compression module. Our training employs Adam optimizer with learning rate 1e-3 and batch size 8, running on 2 NVIDIA H100 GPUs. Our ReVQom configuration uses channel reduction ratio $C_{rr}=16$, $n_q=3$ quantization stages, and codebook sizes $K \in \{4, 16, 64, 256, 1024\}$ to explore the compression-performance trade-off. The EMA decay rate is set to $\alpha=0.8$ based on systematic ablation studies. Loss weights are $\beta_{commit}=0.05$ for VQ commitment loss and $\lambda_{ortho}=0.0001$ for encoder orthogonality regularization. Models are trained for 30 epochs with OneCycle learning rate scheduling. \\
\textbf{Bit Rate Calculation.} The transmission cost per agent is calculated as $R = H \times W \times n_q \times \log_2 K$ bits for ReVQom, compared to $32 \times C \times H \times W$ bits for transmitting uncompressed 32-bit float features \cite{xu2022opencood}, where $H=W=128$ and $C=256$ in our setup. While raw feature transmission appears bandwidth-intensive (8192 bpp), this represents the theoretical baseline that most existing CP methods implicitly assume when transmitting full-precision intermediate features without compression \cite{CodeFilling}. Current collaborative perception approaches focus primarily on \emph{what} features to transmit rather than aggressive compression techniques, making this the appropriate reference for evaluating compression effectiveness.
\subsection{Result and Analysis}
\begin{table}[t]
  \centering
  \caption{Comparison of collaborative perception methods on DAIR-V2X and OPV2V datasets. ReVQom achieves $273\times$-$1365\times$ compression while maintaining competitive performance.}
  \footnotesize
  \begin{tabular}{l|c|c|c|c|c}
  \hline
  Method & $K$ & $bpp$ & AP@0.3 & AP@0.5 & Compr. \\
  \hline
  \multicolumn{6}{c}{\textit{DAIR-V2X Dataset}} \\
  \hline
  No Collaboration & - & 0 & 0.589 & 0.544 & - \\
  F-Cooper \cite{fcooper} & - & 8192 & 0.704 & 0.648 & 1× \\
  V2VNet \cite{wang2020v2vnet} & - & 4096 & 0.695 & 0.635 & 2× \\
  AttFuse \cite{xu2022opencood} & - & 2048 & 0.697 & 0.638 & 4× \\
  CoBEVT \cite{CoBEVT} & - & 8192 & 0.728 & 0.657 & 1× \\
  V2X-ViT \cite{xu2022v2xvit} & - & 6144 & 0.745 & 0.676 & 1.3× \\
  Where2comm \cite{hu2022where2comm} & - & 512 & 0.701 & 0.634 & 16× \\
  \hline
  ReVQom-$\mu$ & 4 & 6 & 0.690 & 0.558 & 1365× \\
  ReVQom-T & 16 & 12 & 0.699 & 0.609 & 683× \\
  ReVQom-S & 64 & 18 & 0.747 & 0.651 & 455× \\
  ReVQom-M & 256 & 24 & 0.753 & 0.666 & 341× \\
  ReVQom-L & 1024 & 30 & 0.725 & 0.636 & 273× \\
  \hline
  \multicolumn{6}{c}{\textit{OPV2V Dataset}} \\
  \hline
  CoBEVT \cite{CoBEVT} & - & 8192 & 0.947 & 0.895 & 1× \\
  ReVQom-S & 64 & 18 & 0.946 & 0.869 & 455× \\
  \hline
  \end{tabular}
  \label{tab:main_results}
\end{table}
\begin{figure}[htb]
  \centering
  \includegraphics[width=\columnwidth]{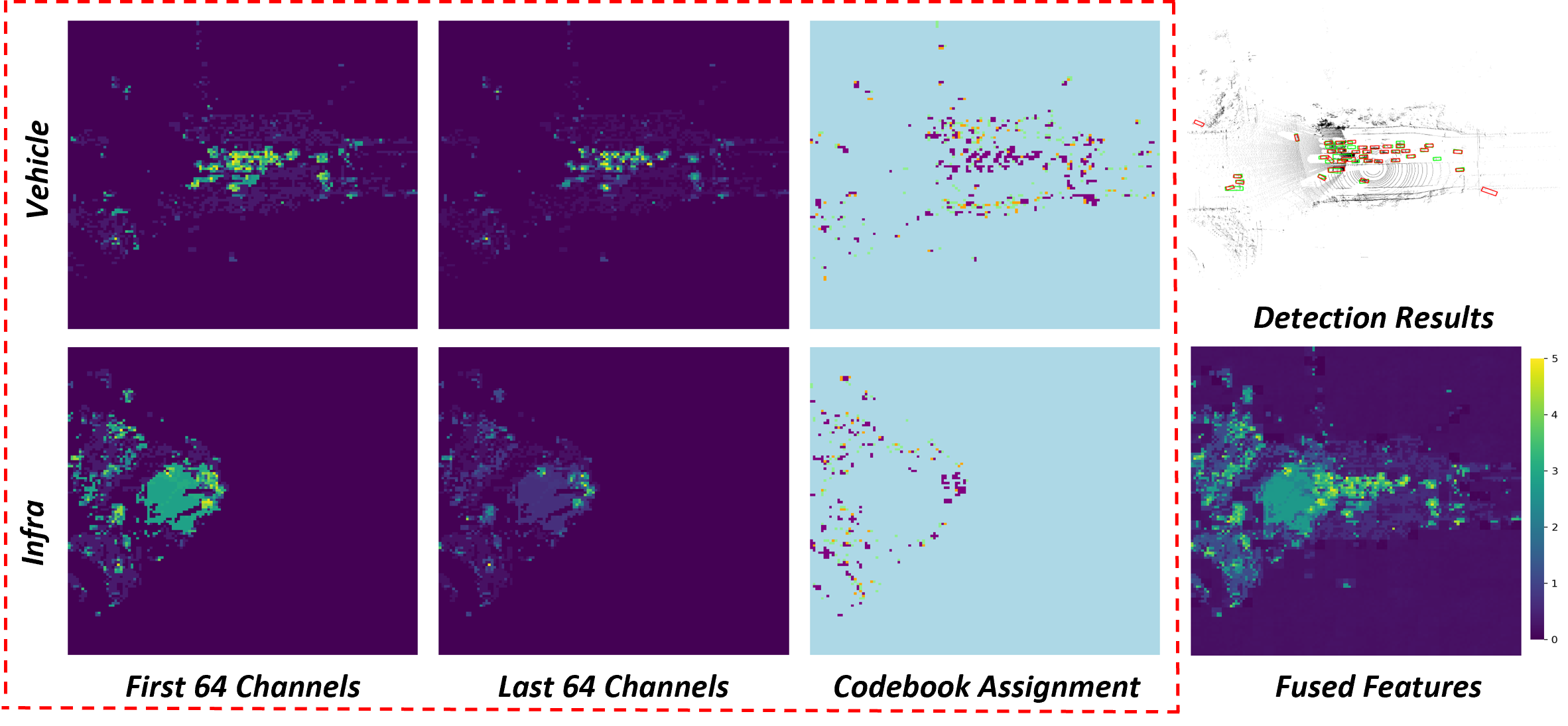}
  \caption{Learned ReVQom codebook assignment visualization ($K=4$ and first quantizer shown for clarity) revealing feature sparsity patterns. The first and last 64 channels show high spatial sparsity with concentrated activations around roads and objects. Codebook assignment (3rd column) demonstrates that Code 0 (background, \textcolor{lightblue}{\rule{8pt}{8pt}}) dominates 96-98\% of spatial locations for both vehicle and infrastructure agents, while Codes 1-3 (\textcolor{lightgreen}{\rule{8pt}{8pt}}, \textcolor{orange}{\rule{8pt}{8pt}}, \textcolor{purple}{\rule{8pt}{8pt}}) efficiently encode semantic foreground regions. This reveals both spatial sparsity (most pixels are background) and channel-wise redundancy across feature maps, enabling ReVQom's aggressive compression while preserving spatial structure for accurate fusion.}
  \label{fig:vis_result}
\end{figure}
\textbf{Quantitative Results.} Table~\ref{tab:main_results} demonstrates ReVQom's excellent compression capabilities across different bit rates. In general, multi-agent collaboration provides $23\%$ AP improvement over single-agent detection. Among baseline methods, V2X-ViT achieves the highest performance ($0.745$ AP@0.3) but at only $1.3\times$ compression compared to uncompressed features. ReVQom-S ($K=64$, $18$ bits) outperforms all baselines including CoBEVT ($0.747$ vs $0.728$ AP@0.3) while achieving $455\times$ compression, suggesting learned compression provides regularization benefits. ReVQom-M attains peak performance ($0.753$ AP@0.3) at $341\times$ compression, surpassing even the best baseline by $1.1\%$. Even ultra-compressed ReVQom-$\mu$ ($1365\times$, $6$ bits) maintains competitive performance ($0.690$ AP@0.3), enabling practical V2X deployment under severe bandwidth constraints. OPV2V evaluation confirms generalization with ReVQom-S achieving $0.946$ AP@0.3 and $0.869$ AP@0.5 at $455\times$ compression, matching uncompressed CoBEVT ($0.947/0.895$) with only $2.9\%$ AP@0.5 drop.\\
\textbf{Qualitative Results.} Figure~\ref{fig:qualitative_results} demonstrates the robustness of ReVQom across different compression levels. At $6$ bits ($K=4$), ReVQom-$\mu$ maintains strong detection performance with most vehicles correctly identified and well-localized, showing only minor differences in bounding box precision compared to higher bit rates. At $18$ bits ($K=64$), ReVQom-S achieves refined detection with improved spatial precision, while $24$ bits ($K=256$) delivers near-optimal performance virtually indistinguishable from raw features. Notably, all compression levels successfully detect the dense vehicle cluster and maintain consistent performance across varying distances, demonstrating ReVQom's ability to preserve critical spatial information even under aggressive compression. \\
\textbf{Codebook Assignment and Usage Statistics.} Figure~\ref{fig:vis_result} reveals semantic structure in learned codebook assignments. The first two columns show the channel wise average feature value for the first and last $64$ channels from vehicle (top) and infrastructure (bottom) sides. Interestingly, the learned codebook assignment coverage for each agent matches well the feature value distribution and BEV detection results. Codebook usage analysis for $K=4$ shows distinct agent-specific patterns. For the scene in Figure~\ref{fig:vis_result}, the vehicle side codebook usage is Code $0$ ($96.3\%$), Code $1$ ($1.1\%$), Code $2$ ($0.7\%$), Code $3$ ($1.9\%$) while infrastructure agent exhibiting even stronger background dominance with Code $0$ ($98.2\%$), Code $1$ ($0.5\%$), Code $2$ ($0.3\%$), Code $3$ ($1.1\%$). Code $0$ primarily encodes background regions in both cases while Codes $1$-$3$ capture semantic objects.
\begin{figure}[htb]
\centering
\includegraphics[width=0.5\textwidth,height=0.15\textheight]{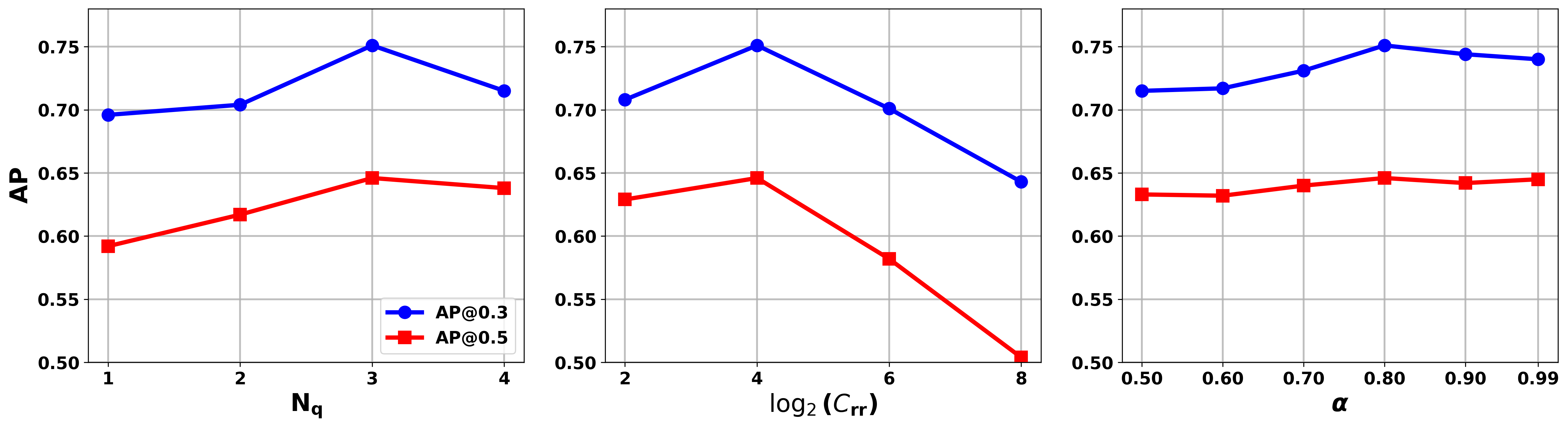}
\caption{Ablation study results for number of quantization stages ($N_q$), channel reduction rate ($C_{rr}$) in logarithmic scale, and EMA decay rate ($\alpha$). $AP@0.3$ and $AP@0.5$ are indicated by (\textcolor{blue}{blue}) and (\textcolor{red}{red}) respectively.}
\label{fig:ablation_groupplot}
\end{figure} \\  
\textbf{Ablation Study.} Figure~\ref{fig:ablation_groupplot} shows systematic ablation results. $\text{EMA}=0.8$ outperforms standard VQ ($\text{EMA}=0.99$), providing faster codebook adaptation suited for dynamic multi-agent scenarios. $N_q=3$ and $C_{rr}=16$ provide optimal performance. Single-stage quantization lacks refinement while $N_q=4$ shows diminishing returns. Channel reduction degrades rapidly beyond $C_{rr}=16$, indicating spatial quantization is more robust than channel compression. Our $K$-sweep reveals $K=64$ provides the best compression-performance trade-off, achieving $99.2\%$ of $K=256$ performance ($0.747$ vs $0.753$ $\text{AP}@0.3$) at $75\%$ bandwidth cost. Notably, $K=1024$ shows performance degradation, demonstrating overfitting in extreme quantization regimes.

\section{CONCLUSION}
\label{sec:conclusion}
We presented ReVQom, a residual vector quantization approach that enables aggressive feature compression for multi-agent collaborative perception while preserving spatial geometry. ReVQom achieves $273\times$--$1365\times$ compression ratios by transmitting only codebook indices instead of full features, maintaining competitive detection performance on real-world V2X scenarios and enabling potential practical deployment of V2X systems.\\
\textbf{Limitations.} Our work focused on LiDAR based datasets, but generalization to other modality (e.g. camera) V2X benchmarks requires further study. Quantization induces feature-space resolution loss, potentially degrading performance at stricter IoU thresholds ($>0.7$). The topics of inaccurate codebook synchronization, packet loss effects and adaptive bitrate selection remain future work.\\
\textbf{Acknowledgment.} This work was supported by US DOT Safety 21 University Transportation Center, Carnegie Mellon University, Pittsburgh, PA, USA.

\vfill\pagebreak
\label{sec:refs}

\bibliographystyle{ieeetr}
\bibliography{refs}

\begin{thebibliography}{10}

\bibitem{Geiger2012CVPR}
A.~Geiger, P.~Lenz, and R.~Urtasun, ``Are we ready for autonomous driving? the {KITTI} vision benchmark suite,'' in {\em Proc. IEEE Conf. Computer Vision and Pattern Recognition}, pp.~3354--3361, 2012.

\bibitem{nuscenes}
H.~Caesar, V.~Bankiti, A.~H. Lang, S.~Vora, V.~E. Liong, Q.~Xu, A.~Krishnan, Y.~Pan, G.~Baldan, and O.~Beijbom, ``nu{S}cenes: A multimodal dataset for autonomous driving,'' in {\em Proc. IEEE Conf. Computer Vision and Pattern Recognition}, pp.~11618--11628, 2020.

\bibitem{waymo2020}
P.~Sun, H.~Kretzschmar, X.~Dotiwalla, A.~Chouard, V.~Patnaik, P.~Tsui, J.~Guo, Y.~Zhou, Y.~Chai, B.~Caine, {\em et~al.}, ``Scalability in perception for autonomous driving: Waymo open dataset,'' in {\em Proc. IEEE Conf. Computer Vision and Pattern Recognition}, pp.~2446--2454, 2020.

\bibitem{wang2020v2vnet}
T.-H. Wang, S.~Manivasagam, M.~Liang, B.~Yang, W.~Zeng, and R.~Urtasun, ``{V2VNet}: Vehicle-to-vehicle communication for joint perception and prediction,'' in {\em Proc. European Conf. Computer Vision}, 2020.

\bibitem{xu2022opencood}
R.~Xu, H.~Xiang, X.~Xia, X.~Han, J.~Li, and J.~Ma, ``{OPV2V}: An open benchmark dataset and fusion pipeline for perception with vehicle-to-vehicle communication,'' in {\em Proc. IEEE Int. Conf. Robotics and Automation}, 2022.

\bibitem{Li_2021_NeurIPS}
Y.~Li, S.~Ren, P.~Wu, S.~Chen, C.~Feng, and W.~Zhang, ``Learning distilled collaboration graph for multi-agent perception,'' in {\em Advances in Neural Information Processing Systems}, 2021.

\bibitem{xu2022v2xvit}
R.~Xu, H.~Xiang, Z.~Tu, X.~Xia, M.-H. Yang, and J.~Ma, ``{V2X-ViT}: Vehicle-to-everything cooperative perception with vision transformer,'' in {\em Proc. European Conf. Computer Vision}, 2022.

\bibitem{CoBEVT}
R.~Xu, Z.~Tu, H.~Xiang, W.~Shao, B.~Zhou, and J.~Ma, ``{CoBEVT}: Cooperative bird's eye view semantic segmentation with sparse transformers,'' {\em arXiv preprint arXiv:2207.02202}, 2022.

\bibitem{cui2022coopernaut}
J.~Cui, H.~Qiu, D.~Chen, P.~Stone, and Y.~Zhu, ``Coopernaut: End-to-end driving with cooperative perception for networked vehicles,'' in {\em Proc. IEEE Conf. Computer Vision and Pattern Recognition}, 2022.

\bibitem{lei2022latencyaware}
Z.~Lei, S.~Chen, Y.~Hu, W.~Zhang, and S.~Chen, ``Latency-aware collaborative perception,'' in {\em Proc. European Conf. Computer Vision}, 2022.

\bibitem{HEAL}
Y.~Lu, Y.~Hu, Y.~Zhong, D.~Wang, S.~Chen, and Y.~Wang, ``An extensible framework for open heterogeneous collaborative perception,'' in {\em The Twelfth International Conference on Learning Representations}, 2024.

\bibitem{xu2023bridging}
R.~Xu, J.~Li, X.~Dong, H.~Yu, and J.~Ma, ``Bridging the domain gap for multi-agent perception,'' in {\em Proc. IEEE Int. Conf. Robotics and Automation}, 2023.

\bibitem{hu2022where2comm}
Y.~Hu, S.~Fang, Z.~Lei, Y.~Zhong, and S.~Chen, ``Where2comm: Communication-efficient collaborative perception via spatial confidence maps,'' in {\em Advances in Neural Information Processing Systems}, 2022.

\bibitem{what2comm}
K.~Yang, D.~Yang, J.~Zhang, H.~Wang, P.~Sun, and L.~Song, ``What2comm: Towards communication-efficient collaborative perception via feature decoupling,'' in {\em Proceedings of the 31st ACM International Conference on Multimedia}, 2023.

\bibitem{CodeFilling}
Y.~Hu, J.~Peng, S.~Liu, J.~Ge, S.~Liu, and S.~Chen, ``Communication-efficient collaborative perception via information filling with codebook,'' in {\em 2024 IEEE / CVF Computer Vision and Pattern Recognition Conference (CVPR)}, 2024.

\bibitem{CPPC}
Z.~Ding, J.~Fu, S.~Liu, H.~Li, S.~Chen, H.~Li, S.~Zhang, and X.~Zhou, ``Point cluster: A compact message unit for communication-efficient collaborative perception,'' in {\em The Thirteenth International Conference on Learning Representations}, 2025.

\bibitem{RVQ}
D.~Lee, C.~Kim, S.~Kim, M.~Cho, and W.-S. Han, ``Autoregressive image generation using residual quantization,'' in {\em Proceedings of the IEEE/CVF Conference on Computer Vision and Pattern Recognition}, 2022.

\bibitem{CoAlign}
Y.~Lu, Q.~Li, B.~Liu, M.~Dianati, C.~Feng, S.~Chen, and Y.~Wang, ``Robust collaborative {3D} object detection in presence of pose errors,'' in {\em Proc. IEEE Int. Conf. Robotics and Automation}, pp.~4812--4818, 2023.

\bibitem{Hetecooper}
C.~Shao, G.~Luo, Q.~Yuan, Y.~Chen, Y.~Liu, K.~Gong, and J.~Li, ``{Hetecooper}: Feature collaboration graph for heterogeneous collaborative perception,'' in {\em Proc. European Conf. Computer Vision}, pp.~162--178, 2024.

\bibitem{GenComm}
J.~Zhou, P.~Dai, Q.~Wei, B.~Liu, X.~Wu, and J.~Wang, ``Pragmatic heterogeneous collaborative perception via generative communication mechanism,'' in {\em Advances in Neural Information Processing Systems}, 2025.

\bibitem{NegoCollab}
C.~Shao, Q.~Yuan, G.~Luo, Y.~Hu, D.~Wang, L.~Yilin, R.~Pan, B.~Chen, and J.~Li, ``{NegoCollab}: A common representation negotiation approach for heterogeneous collaborative perception,'' in {\em Advances in Neural Information Processing Systems}, 2025.

\bibitem{shenkut2024impact}
D.~Shenkut and B.~V.~K. {Vijaya Kumar}, ``Impact of latency and bandwidth limitations on the safety performance of collaborative perception,'' in {\em Proc. Int. Conf. Computer Communications and Networks}, pp.~1--8, 2024.

\bibitem{shenkut2025focalcomm}
D.~Shenkut and V.~Bhagavatula, ``{FocalComm}: Hard instance-aware multi-agent perception,'' {\em arXiv preprint arXiv:2512.13982}, 2025.

\bibitem{vqvae}
A.~van~den Oord, O.~Vinyals, and K.~Kavukcuoglu, ``Neural discrete representation learning,'' {\em Advances in Neural Information Processing Systems}, pp.~6306--6315, 2017.

\bibitem{orthogonal_reg}
N.~Bansal, X.~Chen, and Z.~Wang, ``Can we gain more from orthogonality regularizations in training deep {CNNs}?,'' in {\em Advances in Neural Information Processing Systems}, 2018.

\bibitem{yu2022dairv2x}
H.~Yu, Y.~Luo, M.~Shu, Y.~Huo, Z.~Yang, Y.~Shi, Z.~Guo, H.~Li, X.~Hu, J.~Yuan, and Z.~Nie, ``{DAIR-V2X}: A large-scale dataset for vehicle-infrastructure cooperative {3D} object detection,'' in {\em Proc. IEEE Conf. Computer Vision and Pattern Recognition}, pp.~21361--21370, 2022.

\bibitem{fcooper}
Q.~Chen, X.~Ma, S.~Tang, J.~Guo, Q.~Yang, and S.~Fu, ``{F-Cooper}: Feature based cooperative perception for autonomous vehicle edge computing system using {3D} point clouds,'' in {\em Proc. IEEE/ACM Symposium on Edge Computing}, pp.~88--100, 2019.

\end{thebibliography}

\end{document}